\documentclass[runningheads]{llncs}
\usepackage[T1]{fontenc}
\usepackage{graphicx}
\usepackage{xcolor}

\begin{document}
\title{Removing confounding information from fetal ultrasound images\thanks{Supported by organization x.}}
%
%
\author{***}
\authorrunning{F. Author et al.}
%
\institute{DTU Compute, Technical University of Denmark \and CAMES Rigshospitalet\\
\email{\{kmik, afhar\}@dtu.dk}}

\author{Kamil Mikolaj\inst{1,2} \and
Manxi Lin\inst{1} \and
Zahra Bashir\inst{2} \and
Morten Bo Søndergaard Svendsen\inst{2} \and
Martin Tolsgaard\inst{2} \and
Anders Nymark\inst{1} \and
Aasa Feragen\inst{1}}
\authorrunning{F. Author et al.}
%
%
\maketitle              
\begin{abstract}
Confounding information in the form of text or markings embedded in medical images can severely affect the training of diagnostic deep learning algorithms. However, data collected for clinical purposes often have such markings embedded in them. In dermatology, known examples include drawings or rulers that are overrepresented in images of malignant lesions. In this paper, we encounter text and calipers placed on the images found in national databases containing fetal screening ultrasound scans, which correlate with standard planes to be predicted. In order to utilize the vast amounts of data available in these databases, we develop and validate a series of methods for minimizing the confounding effects of embedded text and calipers on deep learning algorithms designed for ultrasound, using standard plane classification as a test case.

\end{abstract}

\section{Introduction}

\begin{figure}
    \centering
    \includegraphics[width=0.99\linewidth]{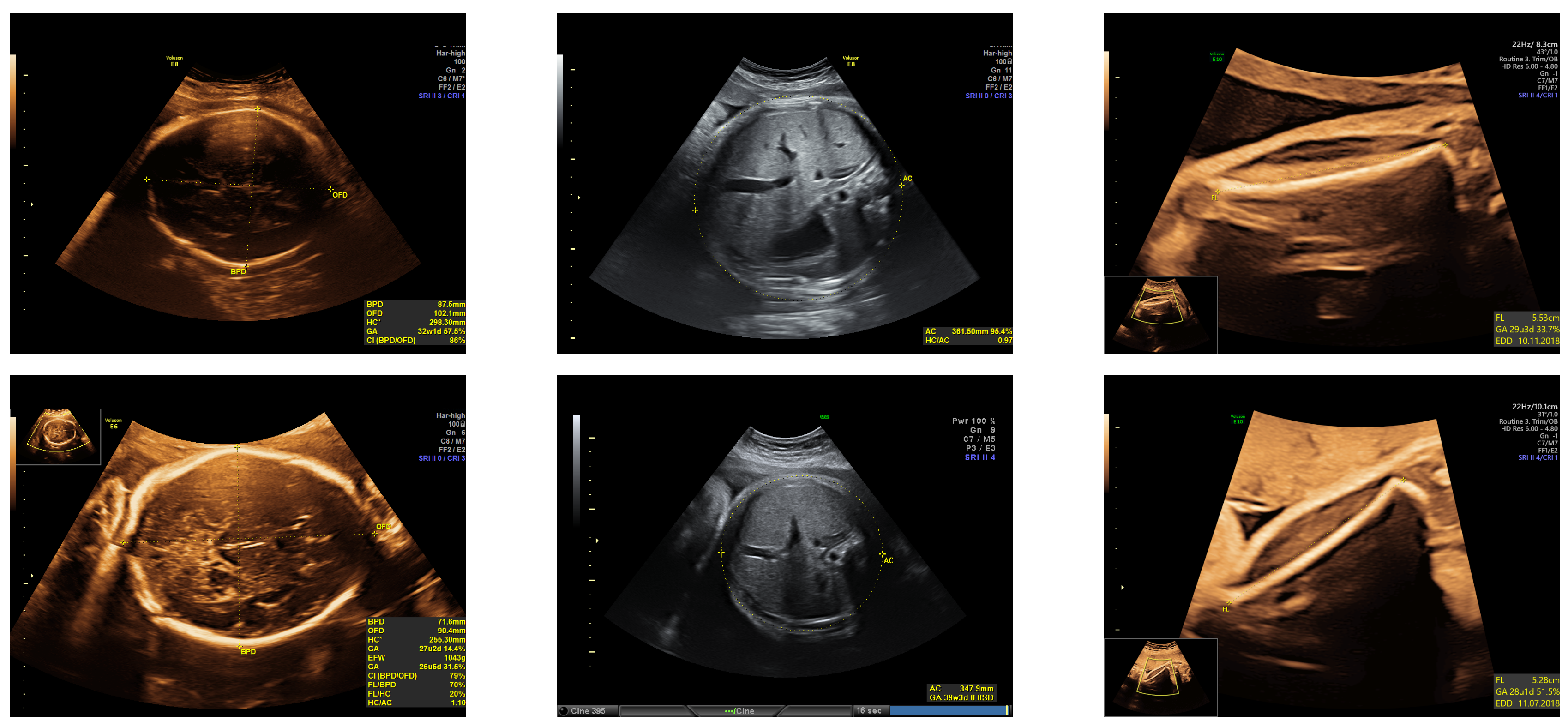}
    \caption{Examples of clinical ultrasound images with text and calipers, i.e.~annotated coordinates for measuring anatomical objects, embedded into the image. Note that both the text labels and the caliper geometry carries information about the particular standard plane that the image contains.}
    \label{fig:confounders}
\end{figure}

Clinical data is a great potential source of data for training medical imaging models: It can come in vast amounts, and represents the nature of data quality encountered in clinical practice. However, when data comes from the clinic, there is little control of the data generating process, and the data may be affected in ways that are suboptimal for training deep learning models. In particular, clinical images sometimes come with embedded text, markings, calipers or other annotations made by the clinician, and these likely carry information correlating with the predictive task at hand. It has recently been shown that markings, stickers and rulers present in dermatological images can confound predictors that aim to diagnose skin lesions~\cite{narla2018automated,winkler2019association,nauta2022uncovering}. In this paper, we consider confounding information present in fetal ultrasound images from clinical screening. As shown in Fig.~\ref{fig:confounders}, these images often have text and calipers embedded in them, which can affect predictors trained on the images. As a case study, we use standard plane classification, which aims to automatically recognize those ultrasound planes required for particular types of screening tests during pregnancy. 

\textbf{Our contribution} First, we quantify the confounding effect of text and calipers embedded in fetal ultrasound images used for training neural networks. Next, we quantitatively assess the success of different methods that aim to remove these confounding effects, ranging from naïve to state of the art text inpainting methods developed for natural images. We show that even simple methods that mask out the confounding information ensure improved generalization to images that do not contain confounding information.

\subsection{Related work}
Recent work in dermatological imaging pointed out that pen markings or rulers that are often present in images of malignant skin lesions from clinical practice, were actually confounding skin lesion diagnosis performed by a CNN approved as a medical device~\cite{winkler2019association,narla2018automated}. More precisely, it was shown that the predictive models performed better on images with rulers and markings than on those without. This discovery spurred research into methods for removing this confounding effect, such as segmenting the lesion as a preprocessing step~\cite{maron2021reducing} to avoid looking at context; inpainting stickers present on benign images to remove them from the training images~\cite{nauta2022uncovering}; inserting prior knowledge into the models~\cite{barnett2021case,rieger2020interpretations}; or adversarially training the neural network to be unable to recognize whether the confounding information is present~\cite{bevan2021skin}.

In our setting, the confounding information is typically embedded into the clinically relevant part of the image. As a result, we cannot apply methods that remove context by segmenting out the object of interest. Instead, we focus on validating both simple and more complex models for removing and inpainting confounding text and calipers.

\section{Methodology}

We assess the confounding effects of embedded text and calipers using standard plane classification for 3rd trimester growth scans as a test case. In order to assess the weight of the fetus, as is commonly done in the 3rd trimester, the clinician needs to obtain standard planes for the head, abdomen and femur. As a typical application is to recognize good standard planes from nonstandard planes that cannot be used for standardized measurements, we also include images that are not standard planes, which should ideally be classified as "Other".

\begin{figure}[t]
    \centering
    \includegraphics[width=\linewidth]{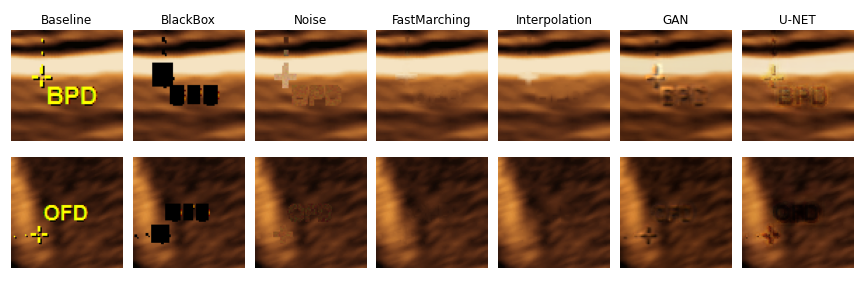}
    \caption{Illustration of inpainting methods replacing both text and calipers on two example image patches.}
    \label{fig:inpainting_comparison}
\end{figure}

We assess six different methods for removing confounding information. Examples of images inpainted with the different methods are found in Fig.~\ref{fig:inpainting_comparison}.

\subsection{Simple methods for removing confounding effects}

The initial four methods consist of first detecting the text and calipers and then replacing them in various ways. 

The text and calipers embedded in the clinically relevant part of the image is (in our training set) always yellow; these are detected via thresholding in hue, saturation, value (HSV) space to segment yellow features. The resulting mask is dilated with a 3x3 structuring element to enlarge the segmentation and connect neighboring elements. 

Additionally, as can be seen in Fig.~\ref{fig:confounders}, most images contain some gray and blue text in the top right corner. To remove this, we first remove everything around the conical ultrasound field of view. A mask is obtained by thresholding the cone in HSV, finding the largest connected component and filling the holes, after which everything else can be masked out. For cases where some blue and grey text is on top of the field of view, the blue text is detected, and everything above is replaced by a black box.

We next consider various approaches to inpainting the yellow masks. 

\subsubsection*{Black box.}
In this first simple approach, the detected yellow mask is overlaid by black boxes spanned by the minimum and maximum $x$- and $y$- coordinates found within every connected component of the mask. 

\subsubsection*{Replacing confounding information by noise.}

As the inpainted black boxes leaves clearly visible information on relative position, caliper geometry etc, we next replace the missing information with noise. For every connected component of the mask, we find its minimal bounding box as above, and expand the bounding box given by the coordinates by 5 pixels in each direction. Next, the contents of the box is replaced by noise as follows: The mean $\mu$ and standard deviation $\sigma$ are computed from the values of those bounding box pixels which are \emph{not} segmented as belonging to text/markings. Those pixels segmented as text/markings are then replaced with noise sampled from a normal distribution $\mathcal{N}(\mu, \sigma/10)$. The scaling of the standard deviation was performed to reduce the visual effect of replacing image pixels with noise; the scaling factor selected from $\{1, \frac{1}{10}, \frac{1}{100}\}$ by optimizing validation set accuracy.


\subsubsection*{Bilinear interpolation.} Third, we apply bilinear interpolation to inpaint the mask.

\subsubsection*{Fast marching inpainting.} Next, we apply a fast marching inpainting method available in OpenCV~\cite{inpainting_telea} to inpaint the mask.

\subsection{Deep learning methods for removing confounding effects}

Next, we assess two different deep learning approaches that aim to generate confounder-free images. Due to the memory cost, the images are divided into $300 \times 240$ pixels patches for model training and inference.  


\subsubsection*{U-Net.} First, we train a U-net~\cite{ronneberger2015u} to generate images without confounding information. We regress the text-free image directly by designing the U-net so that its input and output both have three channels. The pixel intensity of the input images is rescaled to $[-1, 1]$, and the network output is activated by the hyperbolic tangent function. The target labels are images inpainted by bilinear interpolation. We use the SGD optimizer with a learning rate of 0.001 and monitor the mean square root loss. The network is trained for 100 epochs with batch size 8.

\subsubsection*{GAN for generating text-free images.} As an alternative, we train a GAN~\cite{bian2022scene}, which is a state-of-the-art model for scene text removal in natural images. We follow the training settings from the original work. The training set is synthesized by placing the markings detected by thresholding randomly on the images inpainted by bilinear interpolation. This is a common way to construct datasets in scene text removal. 

\subsection{Quantifying the effect of confounders}

In order to quantify confounding effects, we train standard plane classification models to classify standard planes for the growth scans typically performed during the 3rd trimester. For such scans, one typically collects the standard planes for head, abdomen, and femur. 

We train our models first on raw images, and next on images where the confounding text and calipers are removed using the approaches listed above. Both types of models are tested on an internal clinical dataset consisting of images \emph{with} confounders; an internal dataset consisting of images \emph{without} confounders, as well as an external dataset from a different country, where the images do not contain confounders.


\section{Experiments and Results}

\subsection{Standard plane classification models}

We train the Efficientnet B3 architecture~\cite{tan2019efficientnet} for standard plane classification using AdamW with a learning rate of $1e{-4}$, with weighted cross entropy loss to adjust the training to imbalanced data. No augmentation is applied. The classifier is trained for at most 50 epochs, using early stopping with a patience of 5 epochs. We used PyTorch 1.10 for all deep learning based models.

\subsection{Data}

\subsubsection*{Internal database.} We base our evaluation on 3rd trimester growth screening images from the national fetal ultrasound screening programme from ANONYMOUS COUNTRY. The data was collected and used with permission from the ANONYMISED. These images were annotated by an OB-Gyn resident as being either head, abdomen, femur or other, as these are the relevant standard planes for growth estimation. Note, however, that as these images came from the clinical screening database, from a trimester where screening scans are not all made by expert sonographers, they were not all perfect standard planes. To take this into account, the images were given a quality score from 0 (poor) to 10 (excellent), which is shown in Table~\ref{tab:data_quality}.

\subsubsection*{Training data.} 
We performed several experiments with different configurations of the training/validation/test split, all configured with no subject overlap between splits.

We use two different training set configurations for our experiments. All training data is selected from the internal database. The class demographics for both configurations are found in Table~\ref{tab:ds_3trim}.

In the first configuration, the training set contains only images with text and calipers. In the second configuration, the training set contains both images with (77\%)  and without (23\%) text and calipers, sampled in such a way that we have a similar numbers of images per class as in the first configuration.



\begin{table}[]
    \centering
    \begin{tabular}{|c||c|c||c|c||c|c||c|c|c|}
        \hline
        Plane & Train 1 & Val 1 & Train 2 & Val 2 & Local Test  & Local Test  & External Test\\
         & & & & & (with  & (no  & (no \\
         & & & & & confounders) &  confounders) &  confounders)\\
        \hline
        Head & 669 & 147 & 670 & 134 & 127 & 121 & 3092\\
        Abdomen & 786 & 169 & 774 & 171 & 127& 119& 711\\
        Femur & 601 & 118 & 598 & 121 & 127& 113 & 1040\\
        Other & 774 & 181 & 746 & 161 & 42& 38& 4213\\
        N & 2830 & 615 & 2788 & 587 & 423 & 391& 9056\\
        \hline
    \end{tabular}
    \caption{Dataset demographics different experiments, detailing training, validation and test splits from the local and external databases.}
    \label{tab:ds_3trim}
\end{table}

\subsubsection*{Test data}

We test both on data from our own clinical screening database, and on an external dataset from another country~\cite{burgos2020evaluation}.

\begin{table}[b]
    \centering
    \begin{tabular}{|c|c|c|}
    \hline
        Plane & Internal data & Internal data \\
              & (with confounders) & (no confounders) \\
        \hline
        Head &  3.76 $\pm$ 1.74 & 2.92 $\pm$ 2.23\\
        Abdomen & 3.16 $\pm$ 1.63 & 2.46 $\pm$ 1.98 \\
        Femur & 5.13 $\pm$ 1.95 & 4.46 $\pm$ 2.10 \\
    \hline
    \end{tabular}
    \caption{The quality score of given standard planes used in internal data tests. Note the lower quality of the 'no confounders' data.}
    \label{tab:data_quality}
\end{table}

From the internal database we extracted two test sets: One with images that contain text and calipers, and one with images that do not. The test sets were designed to be identically distributed across the classes in order to get comparable performance scores.

Images without text were automatically extracted from the database based on HSV thresholding of the yellow color corresponding to the text. Since the scanner’s model name is also yellow and is placed in the black area outside of the ultrasound field of view, yellow areas that are surrounded by black background are excluded. This is accomplished by morphological dilation of the given area to obtain its neighbouring pixels; if the mean of the neighbours is equal to the background, then it is excluded.

Since the text and calipers are placed by the clinican on those images that are in the end used for predicting clinical outcomes, there is an expected drop in quality for those images that do not contain text and calipers (see Table~\ref{tab:data_quality}).

The external database contains ultrasound scans from 2nd and 3rd trimester, classified into a range of different standard planes, whereas our internal data only contains 3rd trimester. As we use the classes "head", "abdomen" and "femur", the remaining images are given the class "other", which is likely differently distributed than the corresponding images from the internal test sets.





\subsection{Assessing confounding effects}

For each experiment, the training and validation sets were resampled 10 times. Performance was compared to the baseline using t-tests for equality of means of the accuracies, reporting p-values computed over the 10 repeated runs. 


\subsection{Experimental results}

Results for the different training configurations for the standard plane classification are found in Tables~\ref{tab:onlyconf} and \ref{tab:bothconfandnot}.

\begin{table}[]
    \centering
    \begin{tabular}{|c||c|c||c|c||c|c|}
        \hline
        Method & Internal data  & pval & Internal data & pval & External  & pval\\
        & (with  confounders) & &  (no confounders) & & (no confounders) & \\
        \hline
Baseline & \bf 97.0\% $\pm$ 1.1\% & - &  85.6\% $\pm$ 4.1\% & - &  80.5\% $\pm$ 4.0\% & - \\
BlackBox &  96.3\% $\pm$ 1.1\% & 0.17 &  91.7\% $\pm$ 1.4\% & 2.8e-04 &  79.2\% $\pm$ 2.5\% & 0.38 \\
Noise &  95.8\% $\pm$ 1.6\% & 0.070 &  92.9\% $\pm$ 1.1\% & 2.9e-05 &  80.6\% $\pm$ 2.2\% & 0.97 \\
FastMarching &  96.7\% $\pm$ 1.6\% & 0.67 &  93.8\% $\pm$ 1.3\% & 1.0e-05 & \bf 80.8\% $\pm$ 2.6\% & 0.82 \\
Interpolation &  96.1\% $\pm$ 1.5\% & 0.16 &  93.7\% $\pm$ 1.2\% & 1.0e-05 &  80.3\% $\pm$ 2.4\% & 0.91 \\
GAN&  95.9\% $\pm$ 1.7\% & 0.10 &  92.7\% $\pm$ 1.0\% & 4.1e-05 &  80.4\% $\pm$ 3.1\% & 0.96 \\
U-net &  96.5\% $\pm$ 1.3\% & 0.40 & \bf 93.8\% $\pm$ 0.9\% & 7e-06 &  79.4\% $\pm$ 3.0\% & 0.49 \\

        \hline
    \end{tabular} 
    \caption{Classification results training only on images with confounders.}
     \label{tab:onlyconf}
    \end{table}
    
    Note that while there is no significant difference between the methods on the data with confounders, there is a significant difference to the baseline for all methods on the data with no confounders.

    
    
    \begin{table}
    \begin{tabular}{|c||c|c||c|c||c|c|}
        \hline
        Method & Internal data  & pval & Internal data & pval & External  & pval\\
        & (with  confounders) & &  (no confounders) & & (no confounders) & \\
        \hline
        Baseline &  96.6\% $\pm$ 1.0\% & - &  87.4\% $\pm$ 2.9\% & - &  \bf 84.7\% $\pm$ 3.2\% & - \\
        BlackBox &  96.0\% $\pm$ 1.0\% & 0.21 &  92.5\% $\pm$ 1.9\% & 2.1e-04 &  80.9\% $\pm$ 3.1\% & 0.013 \\
        Noise &  95.9\% $\pm$ 1.5\% & 0.23 &  93.5\% $\pm$ 2.3\% & 6.9e-05 &  82.9\% $\pm$ 3.1\% & 0.21 \\
        FastMarching &  96.3\% $\pm$ 1.3\% & 0.59 &  94.5\% $\pm$ 1.8\% & 4.0e-06 & 83.1\% $\pm$ 2.6\% & 0.22\\
        Interpolation &  96.6\% $\pm$ 0.6\% & 1.0 &  94.6\% $\pm$ 1.0\% & 1.0e-06 &  82.8\% $\pm$ 1.8\% & 0.11\\
        GAN &  96.1\% $\pm$ 1.4\% & 0.31 &  93.6\% $\pm$ 1.5\% & 1.3e-05 &  83.0\% $\pm$ 1.5\% & 0.13 \\
        U-net & \bf 96.9\% $\pm$ 1.6\% & 0.70 & \bf 94.7\% $\pm$ 2.1\% & 5e-06 &  81.3\% $\pm$ 2.1\% & 0.011 \\
        \hline
    \end{tabular} 
    \caption{Classification results training both on images with and without confounders.}
    \label{tab:bothconfandnot}
\end{table}


\section{Discussion and conclusion}

We have shown that deep learning algorithms can be confounded when trained on clinical ultrasound images with embedded text or calipers. We have compared several methods for removing text and calipers, ranging from simple detection, removal and classical inpainting or interpolation, to state-of-the-art deep learning models developed to remove text from natural images. All methods have a positive effect by bringing classification performance on clean test images closer to the performance on test images with embedded confounders, even though several of them leave visible artefacts that carry spatial information about the removed confounders. Moreover, the simple methods are performing on par with or even better than the deep learning models. One reason might be that while the deep learning models re-predict the entire image, the simple methods only replace those parts of the image corrupted by text and calipers. Another drawback is that the neural networks are learning texture features for the whole image simultaneously, and inpainting text and calipers with such generic textures may be less beneficial than inpainting with locally inferred texture.

In terms of computational cost, at inference time the deep learning models are about twice as fast as the classical methods. However, the deep learning models also require training time. Moreover, as the classical methods run on CPU, they could likely compete with the inference speed of the deep learning models if they were also implemented on GPU.

We note that the performance on clean images from the internal dataset is still slightly below the performance on images with embedded confounders. This could be due to the generally lower quality of those images that do not have embedded text and calipers. The images without text and calipers are those that were not chosen as standard plane representatives of the clinician -- the highest quality images are the ones used for the clinical calculations and measurements.

\textbf{Why is it so important to be able to train deep learning algorithms on clinical quality images?} Why don't we, instead, perform our own data acquisition obtaining images of the sought quality, but without embedded calipers and text? For standard plane classification, this might be feasible, but would still leave us with far less training data than national screening databases can provide. Even more importantly, however, national screening databases also come with potential for linking with registries cataloguing patient outcomes. Such registries would allow us to train models to recognize rare anomalies and diseases, which we would have no guarantee of finding represented in a smaller dataset acquired for the task. In order to train such models, we need to be able to train our networks robustly, without being affected by confounding information.

\textbf{Why do we try to fix the data, when instead we could try to fix the algorithm?} Indeed, it is desirable to develop algorithms that are fundamentally robust to confounding information. Existing approaches to this problem rely heavily on application specific prior knowledge, such as being able to segment the confounding information away from the image~\cite{barnett2021case}. As, in our case, the confounding information sits right on top of the most relevant part of the image, these approaches do not carry over. By understanding how we might improve training by debiasing our data, we believe we will be better equipped, down the line, to develop algorithms that are inherently robust to confounders and bias.

\subsubsection{Acknowledgements}
The work was partly funded in part by the Innovation Fund Denmark for the project DIREC (9142-00001B); The Capital Region Research Fund and The AI Signature Project, Danish Regions; and the Novo Nordisk Foundation through the Center for Basic Machine Learning Research in Life Science (NNF20OC0062606). The authors acknowledge the Pioneer Centre for AI, DNRF grant P1.

%
%
%
\bibliographystyle{splncs04}
\bibliography{mybibliography}

\end{document}